\documentclass{article}
\usepackage{ijcai18}
\usepackage{amsmath,graphicx,subfigure,booktabs,multirow,epstopdf,setspace}
\usepackage{times}
\usepackage{xcolor}
\usepackage{soul}
\usepackage{array}
\usepackage[utf8]{inputenc}
\usepackage[small]{caption}

\title{Hierarchical Spatial Transformer Network}

\author{
Chang Shu$^{1,2}$,
Xi Chen$^{2}$,
Qiwei Xie$^{2}$,
Hua Han$^{2,3,4}$
\\
$^{1}$ University of Chinese Academy of Sciences, Beijing, China \\
$^{2}$ Institute of Automation, Chinese Academy of Sciences, Beijing, China\\
$^{3}$ School of Future Technology, University of Chinese Academy of Sciences, Beijing, China  \\
$^{4}$ The Center for Excellence in Brain Science and Intelligence Technology, Chinese Academy of Sciences, Shanghai, China\\
\{shuchang2015, xi.chen, qiwei.xie, hua.han\}@ia.ac.cn
}

\begin{document}

\maketitle

\begin{abstract}
Computer vision researchers have been expecting that neural networks have spatial transformation ability to eliminate the interference caused by geometric distortion for a long time.
Emergence of spatial transformer network makes dream come true.
Spatial transformer network and its variants can handle global displacement well, but lack the ability to deal with local spatial variance.
Hence how to achieve a better manner of deformation in the neural network has become a pressing matter of the moment.
To address this issue, we analyze the advantages and disadvantages of approximation theory and optical flow theory, then we combine them to propose a novel way to achieve image deformation and implement it with a hierarchical convolutional neural network.
This new approach solves for a linear deformation along with an optical flow field to model image deformation.
In the experiments of cluttered MNIST handwritten digits classification and image plane alignment, our method outperforms baseline methods by a large margin.
\end{abstract}

\section{Introduction}
Deep learning has achieved great success in the field of computer vision, and has pushed state-of-the-art results forward.
Deep learning encounters a lot of opportunities but also faces many challenges at the same time.
One of challenges is how to make neural networks spatially invariant.

In order to reduce the influence of geometric distortion, one attempt is to design spatially invariant representation.
Handcrafted features like SIFT \cite{Lowe2004Distinctive}, SURF \cite{Bay2008Speeded}, BRISK \cite{Leutenegger2011BRISK} or features learned by CNN \cite{Bruna2012Invariant,Angjoo2014Locally,Sohn2012Learning,Stollenga2014Deep} may obtain spatial invariance in some degree.
However, they may not cover critic features we actually need.
What's more, their emphases are different, thus it is difficult to integrate them in an unified framework.
Deep learning incorporates max-pooling layer to get spatial invariance by only outputting the maximum from a sub-region.
Receptive field of max-pooling layer is relatively small, neural networks won't achieve spatial invariance unless equipped with a plenty number of it.
However, an excess of max-pooling may lose some crucial details of intermediate feature maps.

Data augmentation can increase networks' tolerance on geometric variance through doing spatial transformation to input data.
This approach trades off a sharp increase in the amount of training data for limited spatial invariance.

\begin{figure}[t]
    \centering
    \includegraphics[height=5cm]{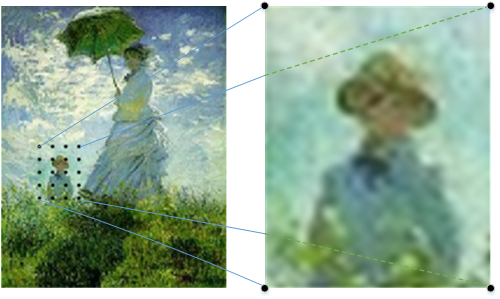}
    \caption{\label{fig:1}An illustration of zoom-in transformation achieved by sampling. Blue lines indicate mappings. Black points represent sampling grid.}
\end{figure}

Above methods don't get to the root of the problem.
Spatial Transformer Network (STN) \cite{jaderberg2015spatial} uses sampling to warp image, as shown in the Figure~\ref{fig:1}.
This approach is differentiable making it possible to be integrated as a layer into a neural network.
STN proposes a brand new way to offer spatial invariance  by providing spatial transformation ability to the network it injects in.
In this case, the network will be able to transform input image to desired pose and shape to avoid negative effects caused by spatial position and geometric distortion.

In the field of image classification where deep learning is widely used, introduction of STN will eliminate partial deformation which would do harm to classification accuracy.
And a state-of-the-art performance has been achieved on the CUB bird dataset \cite{wah2011caltech}.
Furthermore, more and more neural networks \cite{wu2017recursive,zhang2017deep,chang2017clkn,bhagavatula2017faster} apply STN to tackle the task of image alignment, since STN can offer an end-to-end manner to train networks for this issue.

STN can include affine transformation, projective transformation and thin plate spline transformation for application.
These ways of transformation focus on global deformation of an image, but lack the ability to deal with local distortion of details.
Inverse Compositional Spatial Transformer Network (IC-STN) \cite{lin2017inverse} makes improvements on the basis of STN  with the theoretical help of inverse compositional Lucas and Kanada (LK) algorithm.
IC-STN recurrently generates linear transformations using the same module, those linear transformations will work together to warp input image.
The composition of a series of linear transformations is still linear, therefore IC-STN still suffers from inadequate spatial transformation capacity.
Now that STN based methods enables us to achieve spatial transformation within neural networks, how to transform images in a better way becomes a new problem.

To address this issue, we propose hierarchical spatial transformer network (HSTN) to warp image into desired pose and shape.
Firstly, we divide image deformation into two parts: linear deformation and nonlinear deformation. Secondly, we use two modules named linear transformation generator and optical flow field generator to estimate parameters of each part.
Since nonlinear deformation in real scenes is usually hard to be depicted by explicit function, instead we adopt optical flow field to represent it.
Before solving for optical flow field, we need to satisfy optical flow constraint~\cite{Lucas1981An,horn1981determining}, which requires the displacement between the source image and the target image to be small.
To this end, we initialize linear transformation generator with the pre-trained weights from STN, therefore this module will output good initial estimation of linear transformation from the beginning of the training.
Then we jointly train these two modules.
The input of optical flow field generator has been linearly transformed in advance according to the parameters provided by linear transformation generator.
It merely needs to estimate remaining local deformation, thus avoiding the case in which large displacement causes optical flow constraint unsatisfied.
We achieve this process by a convolutional neural network with a hierarchical structure.
Moreover, bending energy penalty~\cite{Galway1990Spline} and smoothness penalty are respectively introduced from approximation theory and optical flow theory to control output of the network.
Finally, we verify our method's effectiveness on the tasks of cluttered MNIST handwritten digits classification and image plane alignment, and our method outperforms baseline methods by a large margin.

Our main contributions of this work can be summarized as follows:
\begin{itemize}
    \item We propose a novel hierarchical convolutional neural network to achieve image deformation.
    \item We put bending energy penalty and smoothness penalty together to control our network's output.
    \item We demonstrate empirically that HSTN's spatial transformation ability is more  powerful than STN and its variants.
\end{itemize}

\section{Related Work}
\begin{figure}[t]
    \centering
    \includegraphics[width=6cm]{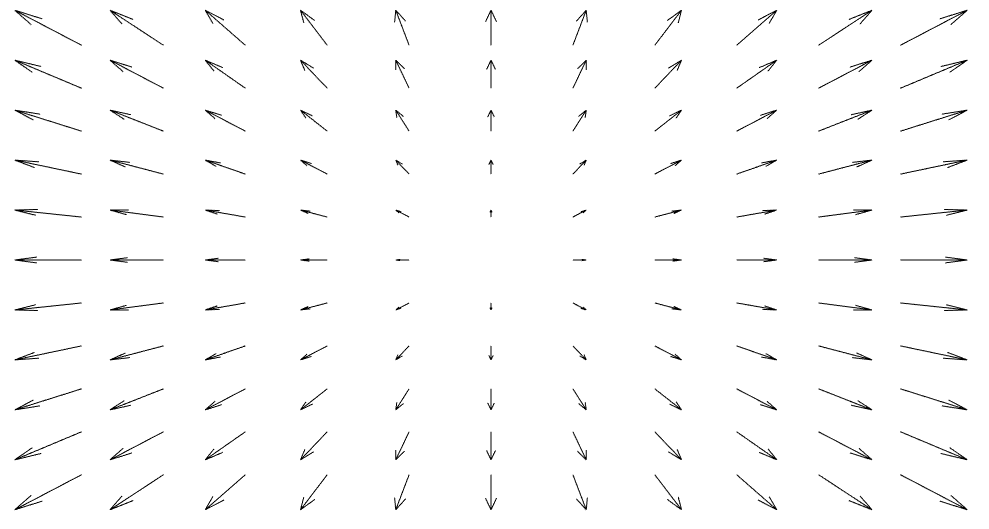}
    \caption{\label{fig:2}An example of the motion field of an image. Arrows denote motional direction and motional distance of each pixel.}
\end{figure}

Determining the deformation of an image is giving the definition of its motion field, namely defining a mapping from a position index $p$ into its motion $\phi (p)$, as shown in Figure~\ref{fig:2}.

In this section, we will review two major categories of theories of modeling image deformation.
One is the approximation theory, the another is the optical flow theory.

\subsection{Approximation Theory}
Approximation theory approximates image deformation by this function: $\phi (p) = c + {\alpha ^T}p + \sum\nolimits_i {{w_i}{f_i}(p)}$.
For clarity of expression, assume that deformation takes place in the  one-dimensional space.
$c + {\alpha ^T}p$ represents linear deformation, where $c$ is a bias term and $\alpha$ is a weighting coefficient.
$\sum\nolimits_i {{w_i}{f_i}(p)}$ approximates nonlinear deformation by a linear combination of a series of basis functions, where ${f_i}$ is a nonlinear basis function and ${w_i}$ is its weighting coefficient.
A basis function is a particular basis for a function space.
Every continuous function in the function space can be represented as a linear combination of basis functions.

\begin{figure}[t]
    \centering
    \includegraphics[height=4cm]{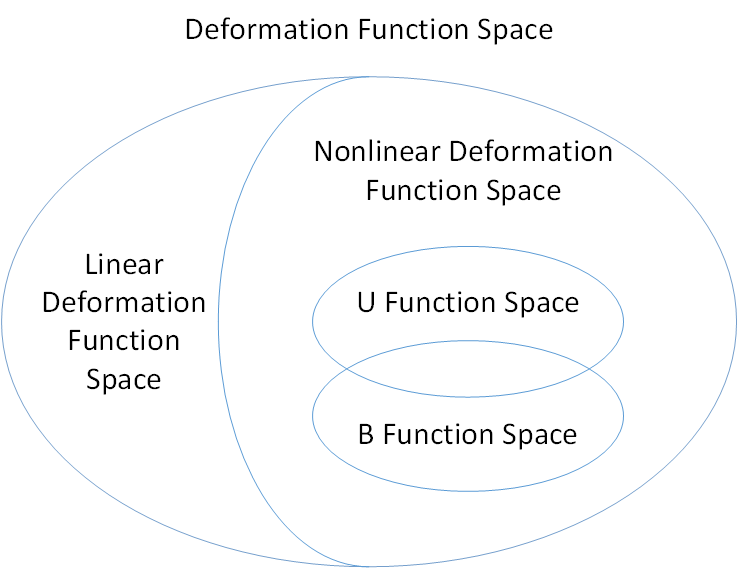}
    \caption{\label{fig:3}An illustration of the structure of deformation function space. }
\end{figure}

Different forms of basis function ${f_i}$ derive different image deformation methods.
When ${f_i}$ is an ``U function'', that is ${{\mathop{\rm U}\nolimits} _i}(p) = {\left( {p - {p_i}} \right)^2}\log {\left( {p - {p_i}} \right)^2}$, corresponding approach is called thin plate spline~\cite{tps}.
When ${f_i}$ is a ``B function'', which is ${{\rm{B}}_i}(p) = C_n^i{(1 - p)^{n - i}}{p^i}$, corresponding approach is called B spline~\cite{bspline}.
Different function spaces spanned by different basis functions are applied to approximate nonlinear deformation function, their relationship can be described by Figure~\ref{fig:3}.

However, these approaches have two disadvantages:
Firstly, the basis functions in use are in a finite number and in a single form, hence method's fitting ability is limited.
Secondly, when nonlinear deformation function is inexplicit or discontinuous, representation by linear combination of basis functions won't be accurate.

\subsection{Optical flow Theory}
Approximation by basis functions is not only way to represent nonlinear deformation.
Optical flow field is often used to estimate motion field.
Optical flow field has great advantage on depicting nonlinear deformation, on account of its free form.
The nonlinear deformation function which is inexplicit or discontinuous can be depicted by optical flow field.

\begin{figure}[t]
    \centering
    \includegraphics[width=8cm]{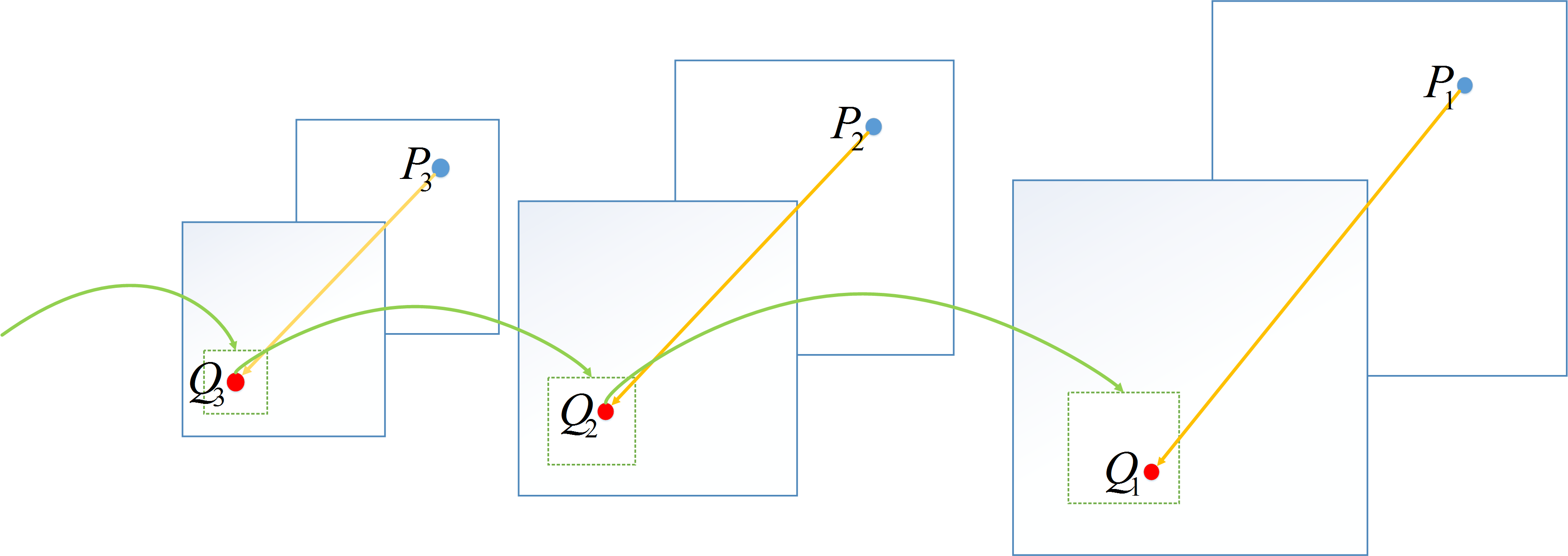}
    \caption{\label{fig:4}An illustration of principle of coarse-to-fine multi-scale iterative scheme. Blue frames denote source images and target images. Blue points and red points are correspondences. Yellow lines denote motions. Green lines represent that results from smaller scale are used for determining possible locations in the larger scale.}
\end{figure}

Optical theory is developed from brightness constancy assumption~\cite{Lucas1981An,horn1981determining}.
Assuming the motion to be small, brightness constancy assumption can be derived into optical flow equation by first-order Taylor series approximation.
Therefore when solving for optical flow field, we have to satisfy optical flow constraint, that is the displacement between source image and target image must be small.
Directly calculating optical flow field will cause large error unless optical flow constraint holds.
Coarse-to-fine multi-scale iterative scheme~\cite{opticflow,siftflow} is a common method to deal with this situation.
Primary idea of coarse-to-fine multi-scale iterative scheme is that resizing image to small enough so that the displacement between source image and target image won't be large.
The optical flow field obtained at smaller scale, will be used as an initial estimation at larger scale.
Then above steps are repeated until iterating to original scale.
This process is summarized in Figure~\ref{fig:4}.

This approach will lose details when resizing images into a small scale.
Error will accumulate to final step and result in inaccurate optical flow.
\section{Hierarchical Spatial Transformer Network}
In this section, we will give definition of our issue, and elaborate on our proposed hierarchical spatial transformer network.

\subsection{Problem Statement}
The deformation of an image can be expressed by a motion field, a motion field of pixels is called an optical flow field.
So solving for motion field of an image can be seen as solving following equation:
\begin{equation}\label{1}
\mathop {\min }\limits_{\bf{w}} {\left| {{\rm{I}}({\bf{x}} + {\bf{w}}) - {\rm{T}}({\bf{x}})} \right|^2}
\end{equation}
Here ${\bf{x}}: = (x,y)$ is a coordinate in an image, ${\bf{w}}: = (u,v)$ is an optical flow we aim to solve.
${\rm{I,T:}}\Omega  \subset {{\bf{R}}^2} \to {\bf{R}}$ denote source image and target image.
Its first-order approximation is known as optical flow equation \cite{horn1981determining}:
\begin{equation}\label{2}
\mathop {\min }\limits_{\bf{w}} {\left| {{\rm{I}}({\bf{x}}) + \nabla {\rm{I}}({\bf{x}}){\bf{w}} - {\rm{T}}({\bf{x}})} \right|^2}
\end{equation}
When value of ${\bf{w}}$ is large, this first-order approximation won't be accurate.
Traditional optical flow methods adopt coarse-to-fine multi-scale iterative scheme to avoid above situation. Nevertheless, as mentioned before, coarse-to-fine multi-scale iterative scheme will do harm to local details and results in inaccurate optical flow field.
So how to make (\ref{2}) work and avoid coarse-to-fine multi-scale iterative scheme's shortcomings at the same time becomes a critical problem.

\subsection{Proposed Method}
\begin{figure*}[t]
    \centering
    \includegraphics[width=17cm]{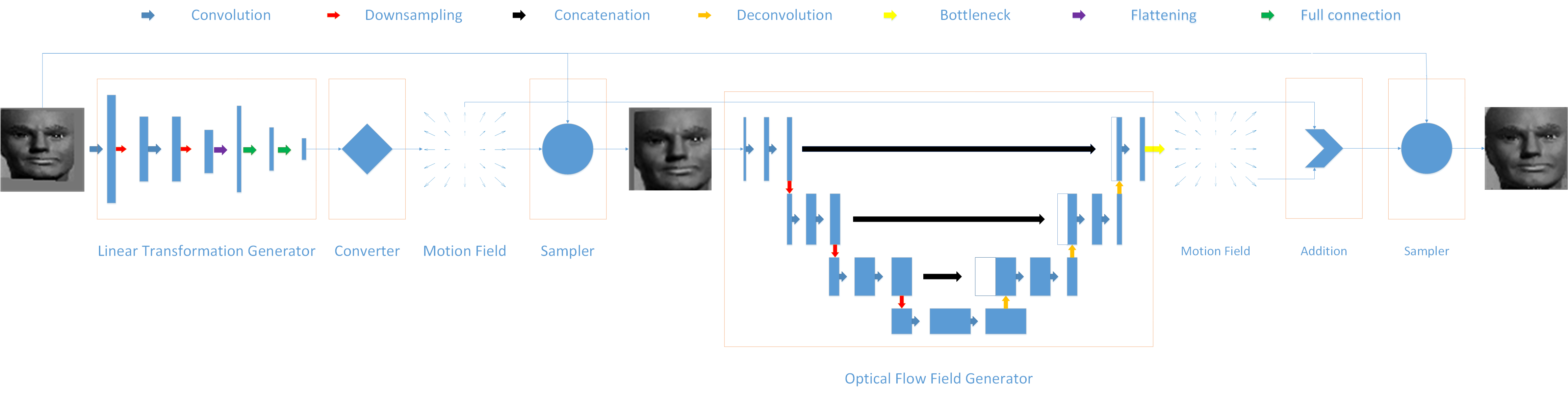}
    \caption{\label{fig:5}The architecture of HSTN.
    It consists of 4 modules: linear transformation, converter, optical flow field generator and sampler.
    The filter sizes of all the convolutional layers are all $3 \times 3$.
    The filter number of all the convolutional layers doubles after each downsampling layer and halves after each deconvolutional layer.
    All the downsampling layers and deconvolutional layers have a stride of 2.
    The bottleneck layer is a convolutional layer with a filter size of $1 \times 1$.}
\end{figure*}

To address these issues, we combine the theory of approximation and the theory of optical flow to propose a novel way to solve for motion field.
Inspired by these two theories, we depict deformation function by a combination of a linear transformation function and an optical flow field.
We first estimate a linear transformation to eliminate large displacement, and then solve for an optical flow field, at this time optical flow constraint will hold.
This process can be mathematically described by following equation:
\begin{equation}\label{3}
\mathop {\min }\limits_{\bf{w}} \mathop {\min }\limits_{\bf{H}} {\left| {{\rm{I}}({\bf{x}} + {\bf{Hx}} + {\bf{w}}) - {\rm{T}}({\bf{x}})} \right|^2}
\end{equation}
Where ${\bf{H}}$ denotes a linear transformation, ${\bf{w}}$ denotes an optical flow field.
${{\rm{I}}({\bf{x}} + {\bf{Hx}} + {\bf{w}})}$ represents ${\rm{I}}$ is linearly transformed and nonlinearly transformed according to ${\bf{H}}$ and ${\bf{w}}$.
The linearised version of (\ref{3}) is:
\begin{equation}\label{4}
\mathop {\min }\limits_{\bf{w}} \mathop {\min }\limits_{\bf{H}} {\left| {{\rm{I}}({\bf{x}} + {\bf{Hx}}){\rm{ + }}\nabla {\rm{I}}({\bf{x}} + {\bf{Hx}}){\bf{w}} - {\rm{T}}({\bf{x}})} \right|^2}
\end{equation}
We first estimate ${\bf{H}}$ to eliminate large displacement, then ${\bf{w}}$ is evaluated when ${\bf{H}}$ is fixed.
In the field of image deformation, nonlinear deformation is very small compared to linear deformation, it mainly exists in local region.
If nonlinear deformation is large, the source image will be warped to be unrecognizable.
Therefore this first-order approximation is feasible after large displacement is eliminated, consequently ${\bf{w}}$ can be accurately solved.
The overall motion field can be described by ${\bf{Hx}}{\rm{ + }}{\bf{w}}$.
${\bf{Hx}}$ stands for global deformation, and ${\bf{w}}$ represents local deformation.
This manner can depict image deformation well.
Above optimization process is achieved by a convolutional neural network with a hierarchical structure, namely, hierarchical spatial transformer network.
The structure of our proposed network is shown in Figure~\ref{fig:5}.

HSTN can be roughly divided into 4 modules: linear transformation generator, converter, sampler and optical flow field generator.
Linear transformation generator takes source image as input and outputs linear transformation parameters.
Converter turns input linear transformation parameters into corresponding motion field and sampler uses it to warp source image.
Linearly transformed image is sent into optical flow field generator.
Optical flow field generator produces an optical flow field.
The sum of obtained optical flow field and the motion field converted from linear transformation is utilized to warp the source image to get final result.
These 4 modules will be explained in detail in the following subsections.

\subsubsection{Linear Transformation Generator}
linear transformation generator takes source image as input, and outputs linear transformation parameters.
Here the linear transformation we use is affine transformation.
Information is extracted from input image by convolution and max-pooling operation.
Through fully connected layer it turns into 6 parameters of affine transformation.
Initializer is used in the last layer of this module to make sure that at beginning of the training, this module outputs an identical transformation.

\subsubsection{Converter}
Converter converts obtained affine transformation parameters into corresponding motion field by using following equation:
\begin{equation}\label{5}
\left[ {\begin{array}{*{20}{c}}
u\\
v
\end{array}} \right] = \left[ {\begin{array}{*{20}{c}}
{a - 1}&b&c\\
d&{e - 1}&f
\end{array}} \right]\left[ {\begin{array}{*{20}{c}}
x\\
y\\
1
\end{array}} \right]
\end{equation}
Where, $a \sim f$ are 6 parameters of an affine transformation, $(x,y)$ is a coordinate in the image, $(u,v)$ is the corresponding motion.

\subsubsection{Sampler}
Sampler do transformation to the input images according to the input motion field, it applies sampling to achieve image deformation like STN~\cite{jaderberg2015spatial}.
We adopt bilinear interpolation to carry out sampling:
\begin{equation}\label{6}
{\rm{T}}({\bf{p}}) = \sum\limits_{i = 1}^4 {{\rm{I}}(\widehat {{{\bf{p}}_i}})\left| {(1,1) - \left| {\widehat {{{\bf{p}}_i}} - {\bf{p}} - {\bf{w}}} \right|} \right|_2^2}
\end{equation}
Where ${\rm T}$ is the output image, ${\rm I}$ is the input image, ${\bf{p}}$ and ${\widehat {{{\bf{p}}_i}}}$ are integer coordinates on the image, and ${\bf{w}}$ is the motion of ${\bf{p}}$, ${\widehat {{{\bf{p}}_i}}}$ is a 4-pixel neighbor (top-left, top-right, bottom-left, bottom-right) of ${{\bf{p}}{\rm{ + }}{\bf{w}}}$.

\begin{figure}[t]
    \centering
    \includegraphics[width=8cm]{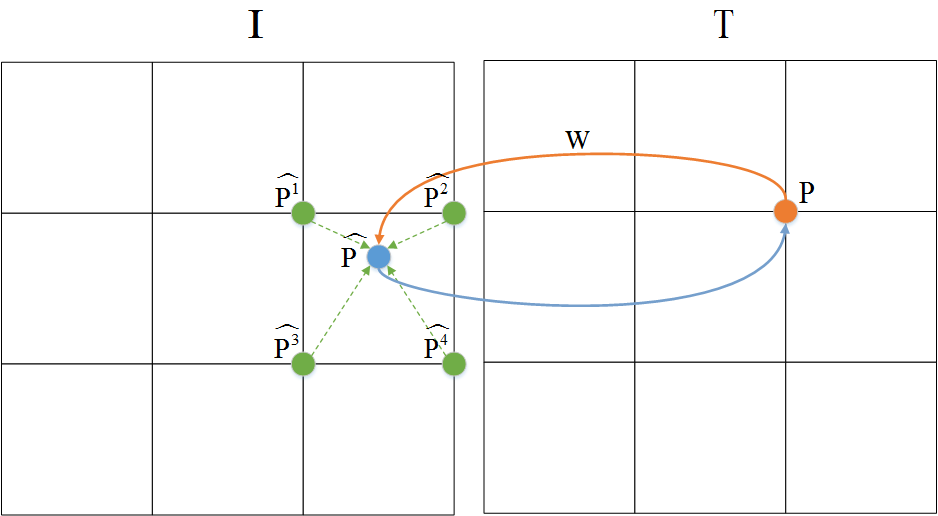}
    \caption{\label{fig:6}The schematic diagram of bilinear sampling by backward mapping. Red point is mapped to blue point. Blue point's pixel value is a linear combination of pixel values of its 4-pixel neighbors (green points), then blue point's pixel value is turned to red point.}
\end{figure}

We achieve bilinear sampling by backward mapping to avoid boundary effect~\cite{lin2017inverse} that the values of some pixels in the output image aren't assigned.
The process of backward mapping is illustrated in Figure~\ref{fig:6}.
Rather than mapping from the input image to the output image, we do the opposite.
Every pixel in the output image will be iteratively mapped into the input image to find its value.

\subsubsection{Optical Flow Field Generator}
Optical flow field generator takes linearly transformed source image as input and output optical flow field for deformation.
It is analogous to an U-net~\cite{ronneberger2015u}, which includes contraction operation and expansion operation to capture critic information and skip-connection to make the most of feature maps from previous layers.
What differs from the U-net is that deconvolutional layers fill in for up-sampling layers.
And initializer is adopted in the last layer of this module to guarantee that values of output optical flow field are all zeros at the beginning of the training.
In addition, we introduce bending energy penalty~\cite{Galway1990Spline} and smoothness penalty respectively from approximation theory and optical flow theory. We add them as regularization terms in the last layer of this module to control the output optical flow $\bf{w}$:
\begin{equation}\label{7}
\ell  = \alpha \left| {\Delta {\bf{w}}} \right|{\rm{ + }}\beta \left| {\nabla {\bf{w}}} \right|
\end{equation}
The first term is the bending energy penalty which penalizes only nonlinear transformation since it gets zero for any linear transformation.
The second term is the smoothness penalty aiming at making optical flow field smoother.
They will be incorporated in the loss function that HSTN optimizes.

\section{Experiments}
In this section, we will describe implement details of compared methods and our proposed method.
MNIST handwritten digits classification and planar image alignment are two classical experiments to test spatial transformation ability of a neural network.
We will report experimental results of these two experiments in following subsections.

\subsection{Cluttered MNIST Classification}
To prove that HSTN can offer better spatial invariance within a classification network, we test methods on the cluttered MNIST handwriting digits database\footnote{$\begin{array}{l}
{\rm{https://s3}}{\rm{.amazonaws}}{\rm{.com/lasagne/recipes/datasets/}}\\
{\rm{mnist\_cluttered\_60x60\_6distortions}}{\rm{.npz}}
\end{array}$}.
This database is modified from classical MNIST handwriting digits database \cite{Lecun2010The}, it is cluttered with more noise.
Implemental details are described below:

(1) \textbf{CNN} is a simple classification network without spatial transformation ability.
Its architecture can be written as
$[conv(32,(3,3)) \to {\rm{P}} \to {\rm{D}}] \times {3} \to dense(256) \to dense(10) \to softmax$.

(2) \textbf{Affine-STN} is a spatial transformer network \cite{jaderberg2015spatial}, which can do affine transformation to input image.
Its architecture can be written as
$[conv(20,(5,5)) \to {\rm{P}}] \times 3 \to dense(50) \to dense(6)$.

(3) \textbf{TPS-STN} is another kind of spatial transformer network \cite{jaderberg2015spatial}, which can do thin plate spline transformation to input image.
Its architecture can be written as
$[conv(32,(3,3))] \times 2 \to {\rm{P}} \to [conv(64,(3,3))] \times 2 \to {\rm{P}} \to [conv(128,(3,3))] \times 2 \to {\rm{P}} \to dense(144) \to dense(32)$.

(4) \textbf{C-STN} is a compositional spatial transformer network \cite{lin2017inverse}, which can do multiple linear transformations to input image.
It has the same structure with \textbf{Affine-STN}.
It uses multiple concatenation of \textbf{C-STN}s to achieve better result, which means each \textbf{C-STN} generates a linear transformation and combination of all those linear transformations is used to transform input image.
A network composed of 2 \textbf{C-STN}s is denoted by \textbf{C-STN-2}, and so on.

(5) \textbf{IC-STN} is an inverse compositional spatial transformer network \cite{lin2017inverse}, which can also do multiple affine transformations to input image.
What different from \textbf{C-STN} is that all the linear transformations are recurrently generated from the same \textbf{C-STN}.
An \textbf{IC-STN} iteratively unfolding 2 linear transformations is denoted by \textbf{IC-STN-2}, and so on.

(6) \textbf{HSTN}'s linear transformation generator is initialized with the pre-trained weights from \textbf{Affine-STN}.
This measure will guarantee that subsequent optical flow field generator gets input without large displacement at the beginning of the training.
And $\alpha$ and $\beta$ in (\ref{7}) are set to 0.01 and 1.

Here $conv(l,(m,n))$ denotes a $l$-filter $m \times n$ convolutional layer,
${\rm{P}}$ denotes a $2 \times 2$ max-pooling layer,
${\rm{D}}$ denotes a dropout layer whose rate is 0.5,
and $dense(n)$ denotes a $n$ unit fully connected layer.
All the methods use \textbf{CNN} as back-end to get classification results.
We use cross entropy as training loss.
We choose Adam \cite{Adam} as optimizer, and default parameters are used.
Learning rate is ${10^{{\rm{ - }}4}}$, and it will be reduced by factor of 10 when valid loss stops decreasing.

\begin{table}[t]
  \centering
  \caption{Comparison of cross entropy and accuracy.}
    \begin{tabular}{p{2.5cm}<{\centering}|p{2.5cm}<{\centering}|p{2.5cm}<{\centering}}
    \hline
    Method & Cross Entropy & Accuracy  \\
    \hline
    CNN          & 0.1560  & 0.9508 \\
    Affine-STN   & 0.0954  & 0.9703 \\
    TPS-STN      & 0.1152  & 0.9772 \\
    C-STN-2      & 0.0714  & 0.9770 \\
    C-STN-4      & 0.0585  & 0.9817 \\
    IC-STN-2     & 0.0813  & 0.9749 \\
    IC-STN-4     & 0.0771  & 0.9766 \\
    \hline
    HSTN          & 0.0540  & 0.9910 \\
    \hline
    \end{tabular}%
  \label{tab:1}%
\end{table}%

Performances of all these methods are illustrated in Table~\ref{tab:1}.
Compared with the CNN method which has no spatial transformation ability, the rest of methods with spatial transformation ability get much higher accuracy.
TPS-STN achieve higher accuracy than Affine-STN for thin plate spline supports nonlinear transformation.
C-STN and IC-STN get higher accuracy than other STN based methods proves that combination of multiple linear transformations can truly improve spatial transformation ability.
Although C-STN achieve better results than IC-STN, IC-STN has lower model capacity which means it has less learnable parameters.
Our proposed HSTN method achieve best result, because it can support more complicate transformation.
We notice that cross entropy isn't completely consistent with accuracy, cross entropy may not be the best option for this task.
Other categorical losses can be adopted, we leaves this for future work.

In the field of classification, stretching and distortion are common factors that influence neural networks' classification efficacy.
Addition of a layer with spatial transformation ability is direct and effective way to deal with this issue.

\subsection{Planar Face Alignment}
\begin{figure}[t]
    \centering
    \includegraphics[height=12cm]{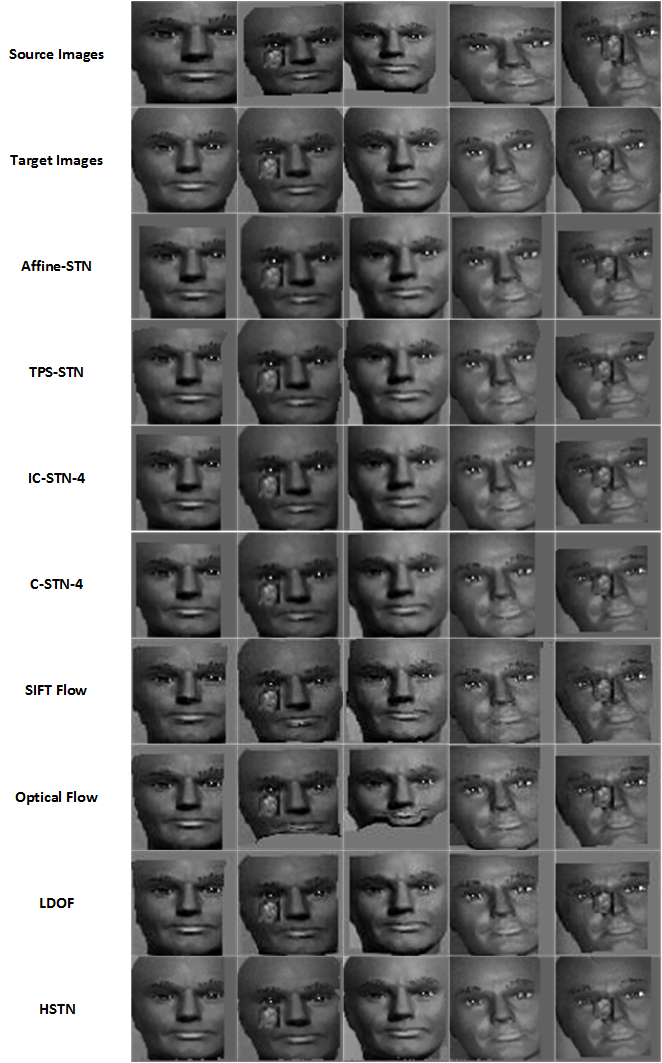}
    \caption{\label{fig:7} An illustration of performance on human face alignment. Source images are perturbed from target images. Methods aim to deform source images to align with target images.}
\end{figure}

In order to exhibit our method's powerful ability to deform an image.
We apply it in the alignment of the images before and after warping.
We use images from a human face database published by \cite{Peng2012RASL}, and we warp them with random linear transformation and elastic deformation, as shown in the first two rows in Figure~\ref{fig:7}.

The same with previous experiment, Affine-STN, TPS-STN, C-STN and IC-STN are used as baselines in this experiment, but they are modified to take concatenated source image and target image as input, and output deformed source image as alignment result.
All these networks optimize their parameters directly by minimizing ${\ell _2}$ error between deformed source image and target image without getting any supervised information, so they are trained in the unsupervised manner.
The rest aforementioned training settings remain the same.
Moreover, three optical flow based methods that use coarse-to-fine multi-scale iterative scheme to satisfy optical flow constraint are also introduced as baselines:

(1) \textbf{Optical Flow} is a classical variational optical flow method which turns optical flow equation into a linear system by using variational approach \cite{opticflow}.

(2) \textbf{SIFT Flow} is derived from optical flow methods using SIFT feature rather than brightness to evaluate alignment accuracy \cite{siftflow}.

(3) \textbf{LDOF} improves on \textbf{Optical Flow}, it offers better initial estimation when solving the linear system.

we use endpoint error (EPE) to measure alignment accuracy, which is Euclidean distance between two images, averaged over all pixels.
In order to avoid interference of background, we only evaluate accuracy on the central part of each image.
Performance of all these methods are shown in the Figure~\ref{fig:7}.
Besides we give quantized evaluation in the Table~\ref{tab:2}.

\begin{table}[t]
  \centering
  \caption{Comparison of EPE and time}
    \begin{tabular}{p{2.5cm}<{\centering}|p{2.5cm}<{\centering}|p{2.5cm}<{\centering}}
    \hline
    Method & EPE & Time \\
    \hline
    Affine-STN   & 6.9781  & 0.0019s \\
    TPS-STN      & 6.4329  & 0.0031s \\
    IC-STN-4     & 6.3330  & 0.0032s \\
    C-STN-4      & 6.3193  & 0.0031s \\
    SIFT Flow    & 8.1266  & 0.5783s \\
    Optical Flow & 6.1780  & 0.0994s \\
    LDOF         & 5.9467  & 0.2572s \\
    \hline
    HSTN          & 1.8088  & 0.0034s \\
    \hline
    \end{tabular}%
  \label{tab:2}%
\end{table}%

In term of speed, neural network based methods are several order of magnitude faster than other methods.
Optical flow method, SIFT flow method and LDOF method all adopt coarse-to-fine multi-scale iterative scheme to deal with large displacement; Nevertheless, deep learning methods directly get transformation parameters through convolution, consequently they take much fewer time.
Given that SIFT flow method computes with SIFT feature rather than brightness, so it takes most time to get alignment result.
Although HSTN has more complicate structure, it gets similar speed with TPS-STN, C-STN and IC-STN.
Affine-STN is faster than other STN based methods, but it sacrifices a lot of accuracy.

In term of accuracy, we can see from Table~\ref{tab:2} that SIFT flow gets most error since this discrete-matching based method can't achieve sub-pixel accuracy.
Compared to STN based methods, optical flow based methods achieve better accuracy since they support nonlinear transformation by estimating optical flow field.
We can see from row 8, column 3 in the Figure~\ref{fig:7} that when encountering large displacement, optical flow method collapses.
The reason is that the optical flow constraint is unsatisfied.
LDOF method can deal with large displacement since it offers good initial estimation of optical flow field before solving optical flow equation.
Our HSTN method outperforms the rest of methods by a large margin, because we use one module to deal with large displacement to satisfy optical flow constraint and use another module to deal with small displacement to make our result more detailed.

Before HSTN, STN based methods can't exceed optical flow based methods, because their spatial transformation ability is limited.
However, HSTN method outperform optical flow based methods by a large margin in term of both accuracy and speed.
And HSTN achieves similar speed and much better accuracy compared with other STN based methods.

Success of deep learning is based on a large amount of labeled data; nevertheless, annotating data is problematic in the field of image alignment.
Above convolutional neural networks are trained in the unsupervised manner, thus they save the trouble of getting ground truth.
This end-to-end framework has broad application prospect in the field of image alignment.
\section{Conclusion}
In this paper, we combine the approximation theory and optical flow theory to propose a novel way to model image deformation and implement it with a hierarchical convolutional neural network.
This way of depiction of image deformation can handle large displacement and small spatial distortion well at the same time.
Our proposed HSTN method achieves superior performance in the experiments of cluttered MNIST handwritten digits classification and planar image alignment.
It can be injected as a layer in any standard neural network for different applications.

\section{Acknowledgements}
This work is jointly supported by Scientific Instrument Developing Project of Chinese Academy of Sciences (No.YZ201671), National Science Foundation of China (NO. 61201050) and Special Program of Beijing Municipal Science and Technology Commission (No.Z161100000216146).

\newpage
\bibliographystyle{named}
\bibliography{ijcai18}

\begin{thebibliography}{}

\bibitem[\protect\citeauthoryear{Bay \bgroup \em et al.\egroup
  }{2008}]{Bay2008Speeded}
Herbert Bay, Andreas Ess, Tinne Tuytelaars, and Luc~Van Gool.
\newblock Speeded-up robust features (surf).
\newblock {\em Computer Vision and Image Understanding}, 110(3):346--359, 2008.

\bibitem[\protect\citeauthoryear{Bhagavatula \bgroup \em et al.\egroup
  }{2017}]{bhagavatula2017faster}
Chandraskehar Bhagavatula, Chenchen Zhu, Khoa Luu, and Marios Savvides.
\newblock Faster than real-time facial alignment: A 3d spatial transformer
  network approach in unconstrained poses.
\newblock {\em International Conference on Computer Vision}, 2, 2017.

\bibitem[\protect\citeauthoryear{Bookstein}{1989}]{tps}
Fred~L. Bookstein.
\newblock Principal warps: Thin-plate splines and the decomposition of
  deformations.
\newblock {\em IEEE Transactions on pattern analysis and machine intelligence},
  11(6):567--585, 1989.

\bibitem[\protect\citeauthoryear{Brox \bgroup \em et al.\egroup
  }{2004}]{opticflow}
Thomas Brox, Andr{\'e}s Bruhn, Nils Papenberg, and Joachim Weickert.
\newblock High accuracy optical flow estimation based on a theory for warping.
\newblock {\em Computer Vision-ECCV 2004}, pages 25--36, 2004.

\bibitem[\protect\citeauthoryear{Bruna and Mallat}{2012}]{Bruna2012Invariant}
J~Bruna and S~Mallat.
\newblock Invariant scattering convolution networks.
\newblock {\em IEEE Transactions on Pattern Analysis and Machine Intelligence},
  35(8):1872--86, 2012.

\bibitem[\protect\citeauthoryear{Chang \bgroup \em et al.\egroup
  }{2017}]{chang2017clkn}
Che-Han Chang, Chun-Nan Chou, and Edward~Y Chang.
\newblock Clkn: Cascaded lucas-kanade networks for image alignment.
\newblock In {\em The IEEE Conference on Computer Vision and Pattern
  Recognition (CVPR)}, 2017.

\bibitem[\protect\citeauthoryear{De~Boor}{1978}]{bspline}
Carl De~Boor.
\newblock A practical guide to splines.
\newblock {\em Applied Mathematical Sciences New York Springer}, Volume
  27(149):157--157(1), 1978.

\bibitem[\protect\citeauthoryear{Galway}{1990}]{Galway1990Spline}
Lionel Galway.
\newblock {\em Spline Models for Observational Data}.
\newblock Society for Industrial and Applied Mathematics,, 1990.

\bibitem[\protect\citeauthoryear{Horn and Schunck}{1981}]{horn1981determining}
Berthold~KP Horn and Brian~G Schunck.
\newblock Determining optical flow.
\newblock {\em Artificial intelligence}, 17(1-3):185--203, 1981.

\bibitem[\protect\citeauthoryear{Jaderberg \bgroup \em et al.\egroup
  }{2015}]{jaderberg2015spatial}
Max Jaderberg, Karen Simonyan, Andrew Zisserman, et~al.
\newblock Spatial transformer networks.
\newblock In {\em Advances in Neural Information Processing Systems}, pages
  2017--2025, 2015.

\bibitem[\protect\citeauthoryear{Kanazawa \bgroup \em et al.\egroup
  }{2014}]{Angjoo2014Locally}
Angjoo Kanazawa, Abhishek Sharma, and David Jacobs.
\newblock Locally scale-invariant convolutional neural networks.
\newblock {\em Computer Science}, 2014.

\bibitem[\protect\citeauthoryear{Kingma and Ba}{2014}]{Adam}
Diederik Kingma and Jimmy Ba.
\newblock Adam: A method for stochastic optimization.
\newblock {\em Computer Science}, 2014.

\bibitem[\protect\citeauthoryear{Lecun and Cortes}{2010}]{Lecun2010The}
Yann Lecun and Corinna Cortes.
\newblock The mnist database of handwritten digits.
\newblock 2010.

\bibitem[\protect\citeauthoryear{Leutenegger \bgroup \em et al.\egroup
  }{2011}]{Leutenegger2011BRISK}
Stefan Leutenegger, Margarita Chli, and Roland~Y. Siegwart.
\newblock Brisk: Binary robust invariant scalable keypoints.
\newblock In {\em International Conference on Computer Vision}, pages
  2548--2555, 2011.

\bibitem[\protect\citeauthoryear{Lin and Lucey}{2017}]{lin2017inverse}
Chen-Hsuan Lin and Simon Lucey.
\newblock Inverse compositional spatial transformer networks.
\newblock 2017.

\bibitem[\protect\citeauthoryear{Liu \bgroup \em et al.\egroup
  }{2011}]{siftflow}
Ce~Liu, Jenny Yuen, and Antonio Torralba.
\newblock Sift flow: Dense correspondence across scenes and its applications.
\newblock {\em IEEE transactions on pattern analysis and machine intelligence},
  33(5):978--994, 2011.

\bibitem[\protect\citeauthoryear{Lowe}{2004}]{Lowe2004Distinctive}
David~G. Lowe.
\newblock Distinctive image features from scale-invariant keypoints.
\newblock {\em International Journal of Computer Vision}, 60(2):91--110, 2004.

\bibitem[\protect\citeauthoryear{Lucas and Kanade}{1981}]{Lucas1981An}
Bruce~D. Lucas and Takeo Kanade.
\newblock An iterative image registration technique with an application to
  stereo vision.
\newblock In {\em International Joint Conference on Artificial Intelligence},
  pages 674--679, 1981.

\bibitem[\protect\citeauthoryear{Peng \bgroup \em et al.\egroup
  }{2012}]{Peng2012RASL}
Yigang Peng, Arvind Ganesh, John Wright, Wenli Xu, and Yi~Ma.
\newblock Rasl: Robust alignment by sparse and low-rank decomposition for
  linearly correlated images.
\newblock {\em IEEE Transactions on Pattern Analysis and Machine Intelligence},
  34(11):2233--2246, 2012.

\bibitem[\protect\citeauthoryear{Ronneberger \bgroup \em et al.\egroup
  }{2015}]{ronneberger2015u}
Olaf Ronneberger, Philipp Fischer, and Thomas Brox.
\newblock U-net: Convolutional networks for biomedical image segmentation.
\newblock In {\em International Conference on Medical Image Computing and
  Computer-Assisted Intervention}, pages 234--241. Springer, 2015.

\bibitem[\protect\citeauthoryear{Sohn and Lee}{2012}]{Sohn2012Learning}
Kihyuk Sohn and Honglak Lee.
\newblock Learning invariant representations with local transformations.
\newblock In {\em International Conference on Machine Learning}, pages
  1339--1346, 2012.

\bibitem[\protect\citeauthoryear{Stollenga \bgroup \em et al.\egroup
  }{2014}]{Stollenga2014Deep}
Marijn~F. Stollenga, Jonathan Masci, Faustino Gomez, and Juergen Schmidhuber.
\newblock Deep networks with internal selective attention through feedback
  connections.
\newblock In {\em International Conference on Neural Information Processing
  Systems}, pages 3545--3553, 2014.

\bibitem[\protect\citeauthoryear{Wah \bgroup \em et al.\egroup
  }{2011}]{wah2011caltech}
Catherine Wah, Steve Branson, Peter Welinder, Pietro Perona, and Serge
  Belongie.
\newblock The caltech-ucsd birds-200-2011 dataset.
\newblock 2011.

\bibitem[\protect\citeauthoryear{Wu \bgroup \em et al.\egroup
  }{2017}]{wu2017recursive}
Wanglong Wu, Meina Kan, Xin Liu, Yi~Yang, Shiguang Shan, and Xilin Chen.
\newblock Recursive spatial transformer (rest) for alignment-free face
  recognition.
\newblock In {\em Proceedings of the IEEE Conference on Computer Vision and
  Pattern Recognition}, pages 3772--3780, 2017.

\bibitem[\protect\citeauthoryear{Zhang and He}{2017}]{zhang2017deep}
Haoyang Zhang and Xuming He.
\newblock Deep free-form deformation network for object-mask registration.
\newblock In {\em Proceedings of the IEEE Conference on Computer Vision and
  Pattern Recognition}, pages 4251--4259, 2017.

\end{thebibliography}

\end{document}